\documentclass[sigconf]{acmart}

\usepackage{multirow}
\usepackage{graphicx}
\usepackage{float}
\usepackage{subfigure}
\usepackage{svg}

\usepackage{algorithm}
\usepackage{algorithmic}
\usepackage{balance}
\usepackage{fancyhdr}

\AtBeginDocument{%
  }

\copyrightyear{2023}
\acmYear{2023}
\setcopyright{acmlicensed}\acmConference[MM '23]{Proceedings of the 31st
ACM International Conference on Multimedia}{October 29-November 3,
2023}{Ottawa, ON, Canada}
\acmBooktitle{Proceedings of the 31st ACM International Conference on
Multimedia (MM '23), October 29-November 3, 2023, Ottawa, ON, Canada}
\acmPrice{15.00}
\acmDOI{10.1145/3581783.3612377}
\acmISBN{979-8-4007-0108-5/23/10}

\begin{document}

\begin{sloppypar}


\title{DFIL: Deepfake Incremental Learning by Exploiting Domain-invariant Forgery Clues}
\author{Kun Pan}
\affiliation{%
  \institution{Zhejiang University}
   \institution{ZJU-Hangzhou Global Scientific and Technological Innovation Center}
     \city{Hangzhou}
  \state{Zhejiang}
  \country{China}
  \postcode{}
}
\email{kunpan@zju.edu.cn}

\author{Yifang Yin}
\affiliation{%
  \institution{Institute for Infocomm Research, A*STAR}
    \city{}
  \state{}
  \country{Singapore}
  \postcode{}
}
\email{yin_yifang@i2r.a-star.edu.sg}

\author{Yao Wei}
\affiliation{%
  \institution{Zhejiang University}
    \city{Hangzhou}
  \state{Zhejiang}
  \country{China}
  \postcode{}
}
\email{weiy@zju.edu.cn}

\author{Feng Lin}
\affiliation{%
  \institution{Zhejiang University}
    \city{Hangzhou}
  \state{Zhejiang}
  \country{China}
  \postcode{}
}
\email{flin@zju.edu.cn}

\author{Zhongjie Ba}
\authornote{The corresponding authors.}
\affiliation{%
  \institution{Zhejiang University}
    \city{Hangzhou}
  \state{Zhejiang}
  \country{China}
  \postcode{}
}
\email{zhongjieba@zju.edu.cn}

\author{Zhenguang Liu}
\authornotemark[1]
\affiliation{%
  \institution{Zhejiang University}
    \city{Hangzhou}
  \state{Zhejiang}
  \country{China}
  \postcode{}
}
\email{liuzhenguang2008@gmail.com}
\author{Zhibo Wang}
\affiliation{%
  \institution{Zhejiang University}
    \city{Hangzhou}
  \state{Zhejiang}
  \country{China}
  \postcode{}
}
\email{zhibowang@zju.edu.cn}

\author{Lorenzo Cavallaro}
\affiliation{%
  \institution{University College London}
  \institution{Zhejiang University}
    \city{London}
  \state{}
  \country{United Kingdom}
  \postcode{}
}
\email{l.cavallaro@ucl.ac.uk}

\author{Kui Ren}
\affiliation{%
  \institution{Zhejiang University}
    \city{Hangzhou}
  \state{Zhejiang}
  \country{China}
  \postcode{}
}
\email{kuiren@zju.edu.cn}
\renewcommand{\shortauthors}{Kun Pan et al.}

\begin{abstract}
The malicious use and widespread dissemination of deepfake pose a significant crisis of trust. Current deepfake detection models can generally recognize forgery images by training on a large dataset. However, the accuracy of detection models degrades significantly on images generated by new deepfake methods due to the difference in data distribution. 
To tackle this issue, we present a novel incremental learning framework that improves the generalization of deepfake detection models by continual learning from a small number of new samples. To cope with different data distributions, we propose to learn a domain-invariant representation based on supervised contrastive learning, preventing overfit to the insufficient new data. To mitigate catastrophic forgetting, we regularize our model in both feature-level and label-level based on a multi-perspective knowledge distillation approach. Finally, we propose to select both central and hard representative samples to update the replay set, which is beneficial for both domain-invariant representation learning and rehearsal-based knowledge preserving. We conduct extensive experiments on four benchmark datasets, obtaining the new state-of-the-art average forgetting rate of 7.01 and average accuracy of 85.49 on FF++, DFDC-P, DFD, and CDF2. Our code is released at \textcolor{blue}{https://github.com/DeepFakeIL/DFIL}.

\end{abstract}



\maketitle
\tabcolsep=4.65pt

\section{Introduction}

Deepfake has swept the globe, inspiring a broad range of inventive and intriguing applications. Today, with a variety of publicly available deepfake tools, anyone can easily produce lifelike fake videos. Unfortunately, the swift spread of malevolent but convincing fake videos has resulted in a significant crisis of trust due to the substantial amount of confusion and deception they have caused.
The technology came into the limelight when unscrupulous people maliciously produced pornographic videos of Hollywood stars~\cite{deepfakenews1} and a variety of fake news~\cite{deepfakenews2}. 
These facts suggest that the security problems caused by deepfake technology should not be underestimated. However, the rapid emergence of new deepfake methods has made forgery detection extremely challenging. Thus, it is critical and urgent to develop a detection technology that can cope with various potential forgery methods.

Existing forgery detection methods mostly focus on using deep learning to find forgery traces in forged images. Based on the observations of forged images, early work proposes to artificially locate forgery traces~\cite{li2020face}. For example, in the face replacement detection method, researchers found that the forgery traces usually exist in the inconsistency of the head pose~\cite{yang2019exposing}, the eyes~\cite{li2018ictu} and teeth level details~\cite{haliassos2021lips}. However, due to the proliferation of forgery techniques, these traces are mostly hidden or eliminated, resulting in these methods performing unsatisfactorily nowadays. Recent research learns from a large collection of forged images and real images to detect more general and subtle forgery traces~\cite{zhuang2022uia,liu2021spatial,shiohara2022detecting}. However, such methods still suffer from performance degradation when applied to images generated by new forgery methods. 

To address this issue, we present to our knowledge the first domain-incremental learning framework for deepfake detection, which can quickly generalize the deepfake detection models to a new domain based on a few new samples. Different from traditional continual learning problems, a domain in deepfake detection is not only defined by the real data but also the fake data generated by various and rapidly emerging forgery methods. 
This poses new challenges in deepfake continual learning as 1) we are not able to collect enough samples for every new task (\emph{i.e.}, domain) in practice, and 2) learning from a large number of tasks sequentially leads to a more serious data imbalance issue between the new and the already learned domains. 
Existing continual learning methods rely more on the data of the new task, while mitigating the catastrophic forgetting issue on previous tasks based on a small number of past samples. However, due to the data limitation in deepfakes, we cannot rely on the small number of new task samples to update the feature distribution. Thus, it is critical for us to model the relations between the new and the old data to prevent our model from overfitting to the limited new task samples.

\begin{figure}[t] 
\centering 
\includegraphics[width=0.45\textwidth]{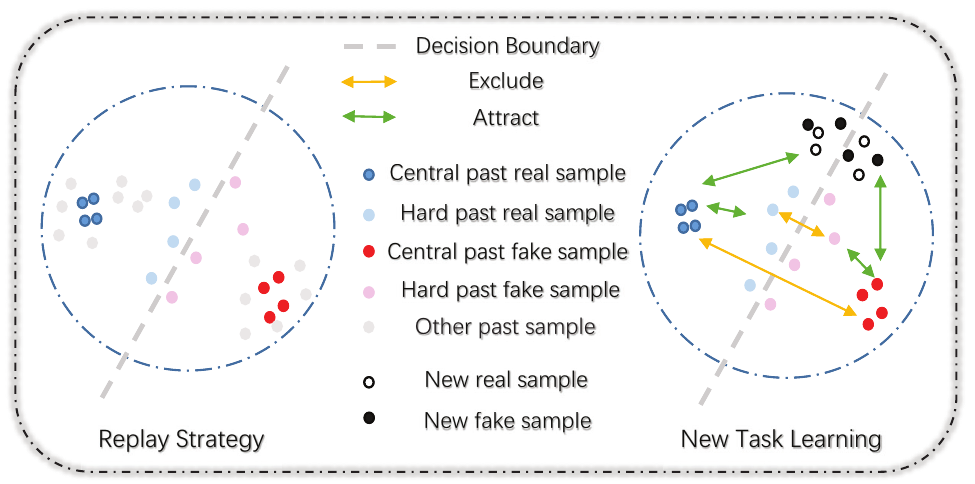} 
\caption{Illustration of our proposed replay strategy and domain-invariant representation learning.} 
\label{fig:intro} 
\end{figure}

The above observations motivate us to think if it is possible to learn a domain generalized representation during continual learning, such that the catastrophic forgetting on previous tasks and the data limitation on new tasks can be solved in one unified framework. To this end, we propose to learn domain-invariant representations based on supervised contrastive learning \cite{khosla2020supervised} by exploiting the semantic relations among sample-to-sample pairs from different domains. As illustrated in Figure~\ref{fig:intro}, we create a replay set consisting of representative old task samples, which is merged with the limited new task samples to perform supervised contrastive learning. Thereby, semantically-consistent representations are learned across domains by pulling together samples of the same class and pushing apart samples of different classes in the shared semantic space. The creation of the replay set is a key component in our framework. Its impacts are twofold. First, it prevents forgetting past knowledge by replaying past samples. Second, it promotes domain-invariant representations by aligning with new samples. 
Here we propose to select both hard samples based on information entropy and central samples based on feature distance to class centroids to form a representative replay set. Our replay strategy shows significant benefits in both aspects, outperforming existing methods consistently in deepfake detection continual learning.
To further mitigate catastrophic forgetting, we adopt the multi-perspective knowledge distillation~\cite{hinton2015distilling} to regularize our model based on both intermediate features and soft labels.
We conduct extensive experiments on four benchmark ceepfakes detection datasets, namely FF++ \cite{rossler2019faceforensics++}, DFDCP \cite{dolhansky2019deepfake}, CDF2 \cite{li2020celeb}, and DFD \cite{rossler2019faceforensics++}. 
The experimental results show that our proposed method can effectively detect fake samples in the new task by learning from only a few new samples. Here we summarize our contributions as follows:

\begin{itemize}
\item We present a novel model-agnostic, domain-incremental learning framework for deepfake detection, which can effectively detect forged samples in a new domain based on only a few new samples.

\item We propose to learn semantically-consistent representations across domains based on supervised contrastive learning. A new replay strategy is introduced to prevent catastrophic forgetting and to promote domain-invariant representation learning simultaneously.

\item We conduct extensive experiments on four benchmark datasets. The experimental results show that our proposed method outperforms existing incremental learning and deepfake detection methods by a significant margin.

\end{itemize}

\section{Relate Work}

\textbf{Deepfakes Generation.}
Inspired by the great success of deep learning in computer vision~\cite{MM2017,CVPR2021DualConsecutiveNetwork,CVPR2019MotionPrediction,yang2023action,zhang2021token,hao2022group,deng2023counterfactual,dong2018predicting,dong2022partially,yin2021enhanced,yin2021gps2vec,yang2022video,yang2020tree,yin2022mix}, deepfake generation technologies \cite{korshunova2017fast,FaceSwapDevs2018,thies2016face2face,thies2019deferred,nirkin2019fsgan,nirkin2019fsgan,li2019faceshifter,chen2020simswap,wang2021hififace,gao2021information,DeepfakeWholebody,ba2023transferring} have also been emerging rapidly in recent years. Fast Face-swap \cite{korshunova2017fast} uses CNNs to extract face information and style information for face forgery. FaceSwap \cite{FaceSwapDevs2018} uses two Encoder-Decoders to achieve facial forgery. Face2Face \cite{thies2016face2face} proposes a real-time highly realistic facial expression transmission from source to target video. NeuralTextures \cite{thies2019deferred} shows facial reenactment by using a NeuralTextures-based rendering approach. FSGAN \cite{nirkin2019fsgan} provides a Many2Many face swapping paradigm to make the creation of forged images more efficient and simple. FaceShifter \cite{li2019faceshifter} solves the impact of the hair masking problem on image quality. SimSwap \cite{chen2020simswap} makes forgery results realistic nature by injecting identity information. HifiFace \cite{wang2021hififace} mainly introduces the geometric structure signal of 3DMM to make the generated swapped face shape consistent with the sourced face. 
Moreover, many forgery tools have been open-sourced, such as FakeApp \cite{clark2019adversarial} and ZAO \cite{ZAOAPP2020}, which makes forgery techniques more accessible to the public. 

\textbf{Deepfakes Detection.}
People are facing the threat of forgery due to the abuse of deepfake. Thus, the detection of deepfake has become an urgent yet challenging task. A significant number of deepfakes detection methods have been proposed recently~\cite{dong2022protecting,li2020face,shiohara2022detecting,chen2022self,haliassos2021lips,zhao2021multi,frank2020leveraging,li2021frequency,liu2021spatial,luo2021generalizing,qian2020thinking} proposed for deepfakes detection. Face X-ray~\cite{li2020face} and SBI~\cite{shiohara2022detecting} create extra forgery images to train the model, improving the model detection performance. LFA~\cite{frank2020leveraging} believes that all generative model must go through the up-sampling process, and the frequency domain of the up-sampling image will be significantly different from the real image, so direct recognition in the frequency domain will achieve very good results.
Though promising results have been obtained, challenges persist in the model's generalization ability to the constantly emerging new deepfake algorithms. Without the capability of adapting to new forgery algorithms, a detection model will soon become infeasible in real-world applications. 

\textbf{Incremental Learning.}
Incremental learning methods can be roughly divided into three categories, namely model structure-based method \cite{rusu2016progressive,mallya2018packnet,singh2021rectification,yan2021dynamically,zhang2021few}, regularization-based method \cite{li2017learning,kirkpatrick2017overcoming,douillard2020podnet,zenke2017continual,pan2020continual}, and replay-based method \cite{rebuffi2017icarl,lopez2017gradient,bang2021rainbow,mai2021supervised,yu2020semantic,shin2017continual,chaudhry2019tiny}. 
Existing incremental learning methods mostly train with sufficient new task data, while focusing on proposing cost-efficient algorithms to preserve past knowledge. However, as aforementioned, the limitations of data availability and data imbalance in a practical forgery detection application pose new challenges for deepfakes continual learning. We thus focus on these new challenges and investigate the possibility of learning with a few new samples only.

\begin{figure*}[t] 
\centering 
\includegraphics[width=1\textwidth]{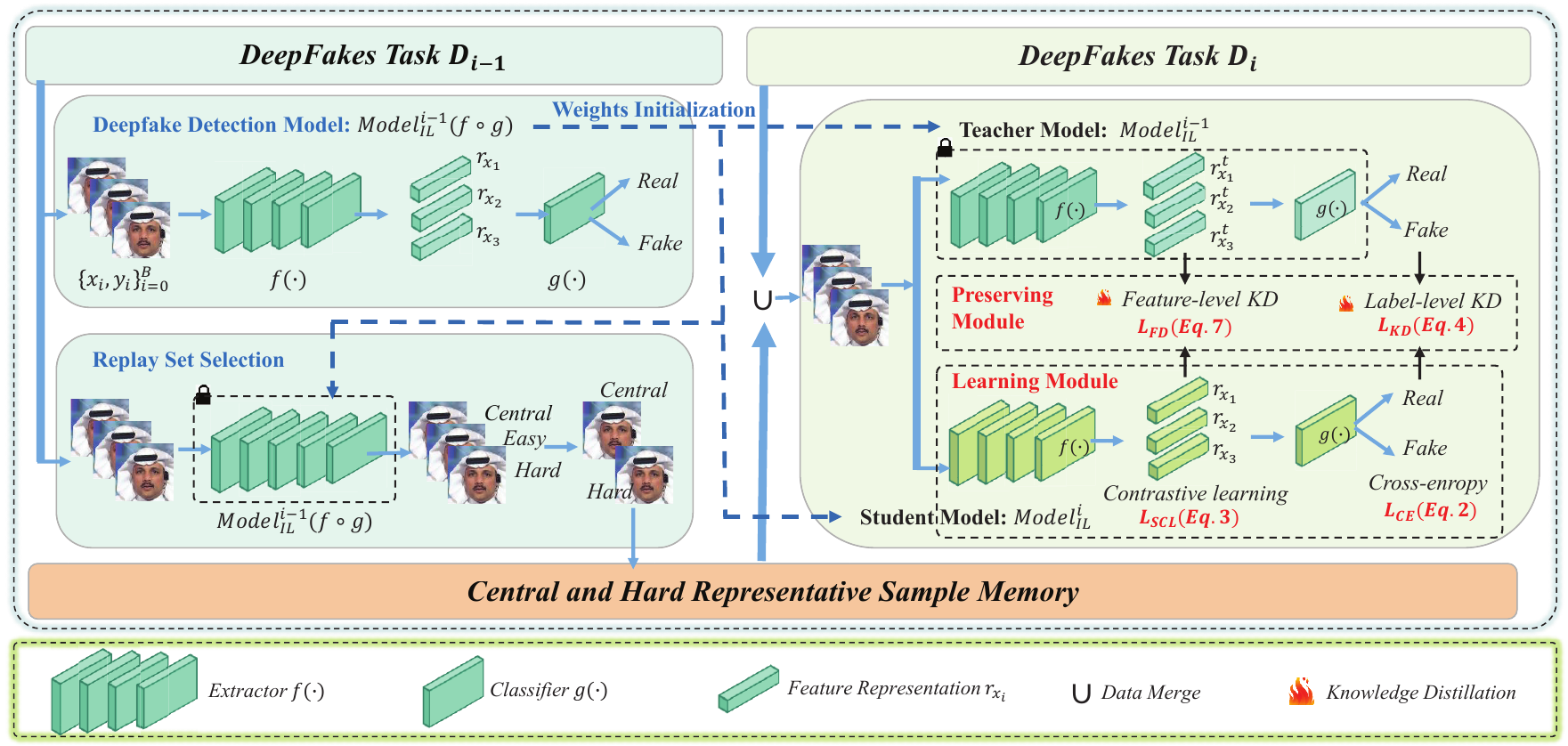} 
\caption{The architecture of our method. Specific modules show the training process for Task $i-1$ and Task $i$. The objection functions of Task 1 and Task 2 are different, and there is no multi-perspective knowledge distillation Loss in Task 1.} 
\label{fig:architecture} 
\end{figure*}

\textbf{Contrastive Learning.}
The main idea of contrastive learning is to attract similar samples while repel different samples \cite{chen2020simple,he2020momentum,wu2018unsupervised,khosla2020supervised}. Recently, some work has used contrastive learning to detect deepfake \cite{fung2021deepfakeucl,xu2022supervised,sun2022dual}. DeepfakeUCL \cite{fung2021deepfakeucl} proposes an unsupervised contrastive learning method for deep detection. Consup \cite{xu2022supervised} uses supervised contrastive learning for deepfakes detection. DCL \cite{sun2022dual} learns generalized feature representation by designing contrastive learning at multi-granularity. In this paper, the purpose of using contrastive learning is to constrain the positions of real and fake images in the feature space, thereby facilitating the learning of new tasks for the detection model.

\section{Problem Formulation and Framework Overview}
\subsection{Problem Formulation}
Whenever a new forgery method emerges, it is desirable that a detection model can be refined based on a small number of new samples to cope with the new forgery method. We formulate this problem as a deepfake incremental learning task. A continuous set of deepfake tasks is formally given as $D_C = \left \{D_1,D_2,..., D_N  \right \} $, where $D_i$ denotes the $i$-th deepfake detection task containing both real and fake images, and $N$ is the total number of tasks. In the real scenario, the first dataset we use to train the detection model should be the union of all the existing forgery methods, while the subsequent datasets are collected on-the-go when new forgery algorithms appear. Therefore, the number of images in the first dataset $|D_1|$ is much larger than the number of images in the subsequent datasets $|D_N|_{N\ne 1}$. We take out the forged datasets sequentially from $D_C$ to train the detection model incrementally. Let $i$ denote the $i$-th deepfake detection task. The model before and after training with $D_i$ are denoted as $Model^{(i-1)}_{IL}$ (the teacher model) and $Model^{(i)}_{IL}$ (the student model), respectively. 
Deepfake incremental learning mainly studies the problem of updating $Model^{(i-1)}_{IL} \rightarrow Model^{(i)}_{IL}$ to cope with a new task $D_i$ ($i\geq 2)$ based on the previous model $Model^{(i-1)}_{IL}$, the current task $D_i$, and previous tasks. For $D_1$, as no past knowledge is to be preserved, the problem can be simplified to a normal deepfake detection task on a single dataset.


Next, we will briefly introduce the overview of the proposed framework (Section~\ref{sec:overview}), followed by the technical details of our proposed deepfake incremental learning on $D_i$ (Sections~\ref{sec:learning},~\ref{sec:preserving}, and~\ref{sec:replay}).
Finally, an overall deepfake incremental learning procedure on $D_C= \left \{D_1,D_2,..., D_N  \right \}$ will be illustrated in Section~\ref{sec:algorithm}.

\subsection{Framework Overview}
\label{sec:overview}
Figure~\ref{fig:architecture} illustrates the overall architecture of our proposed deepfake incremental learning on task $D_i$. It mainly consists of a deepfake detection model, a replay set selection module, a domain-invariant representation learning module, and a past model's knowledge preserving module. For generality, we define the deepfake detection model to be composed of a feature extractor network $f(\cdot)$ and a linear classification network $g(\cdot)$~\cite{he2016deep,rossler2019faceforensics++}. A data sample is given as {$(x_j,y_j ) \in D_i=\left \{ (x_0,y_0), (x_1,y_1 ),\dots,(x_n,y_n )\right \}_{n=|D_i |}$} where $x_j$ is an image, $y_j$ is the label of the image, and $D_i$ is the $i$-th deepfake dataset in $D_C$. By passing $x_i$ to the feature extractor network $f(\cdot)$, we can obtain a feature representation $r_{x_j} = f(x_j)$. Then, $r_{x_j}$ will be passed to the linear classification network $g(\cdot)$ to generate a logit score $z_{x_j} = g(r_{x_j})$. Finally, $z_{x_j}$ goes through a softmax layer to obtain the probabilistic real-fake binary classification result $p_{x_j}$. It is worth mentioning that our proposed method is model-agnostic. Our method can be applied to any deepfake detection model where consistent performance gain can be obtained. 

\section{DFIL : Deepfake Incremental Learning}
Here, we propose a deepfake incremental learning framework, denoted as \textbf{DFIL} (\textbf{D}eep\textbf{F}ake \textbf{I}ncremental \textbf{L}earning). For a new task $D_i$ where $i\geq 2$, DFIL aims to (1) learn semantically-consistent representations across old and new tasks through cross-entropy loss ($L_{CE}$) and supervised contrastive learning ($L_{SCL}$), and (2) preserve the past model's knowledge through multi-perspective knowledge distillation ($L_{KD}$ and $L_{FD}$). As shown in Figure~\ref{fig:architecture}, the overall loss function of DFIL is given as
 \begin{equation}
L_{DFIL} = \underbrace{L_{CE}+ \alpha L_{SCL}}_{(1)Learning}  + \underbrace{\beta L_{KD} + \gamma L_{FD}}_{(2)Preserving} 
 \end{equation}

For optimization, we draw each mini-batch from the union of current task samples and representative previous samples. The set of representative previous samples is usually referred to as the replay set, the selection of which will be introduced in Section~\ref{sec:replay}. For presentation simplicity, we will omit the task indicator $(i)$ as the incremental learning procedure is the same for every $D_i$ ($i\geq 2$). In the rest of this paper, we will use teacher model and student model to refer to $Model^{(i-1)}_{IL}$ and $Model^{(i)}_{IL}$, respectively. 

\subsection{Learning Domain-invariant Representations}
\label{sec:learning}
Recall that in deepfake continual learning, the number of new task samples is usually small, which cannot well represent the new task data distribution. We thus propose to align the feature representations between the old and new tasks based on supervised contrastive learning, which is beneficial for both new task learning and previous knowledge preservation. 

\textbf{Cross-Entropy Loss.} The student model can directly learn from the task labels by the cross-entropy loss ($L_{CE}$) so that the feature distribution of the real and fake samples can be separated. The cross-entropy loss is computed as:
\begin{equation}
L_{CE} =  \sum _{i=1}^B -[y_i\cdot log(p_{x_i(0)}) + (1-y_i)\cdot log(1-p_{x_i(0)})]
\label{eq:LCE}
\end{equation}
where $y_i$ is the label of sample $x_i$, 0 for real class and 1 for fake class, $B$ is batch size, and $p_{x_i(0)}$ is the probability that sample $x_i$ is predicted to be the real class.


\textbf{Supervised Contrastive Loss} 
The cross-entropy loss enables the model to identify fake samples. However, it does not align the samples in the feature distribution level across tasks. Intuitively, we would like to pull positive sample pairs (\emph{i.e.,} samples of the same class) from different domains closer while pushing other negative pairs (\emph{i.e.,} samples of different classes) further apart. We thus utilize the supervised contrastive loss \cite{khosla2020supervised} to generate a shared feature space across tasks. 
One biggest difference between supervised contrastive learning and self-supervised contrastive learning is that the former leverages task labels so positive samples for each anchor point is more informative. In the self-supervised contrastive learning, the positive samples for each anchor point are formed by the augmented views of its own, while in the supervised contrastive learning, the positive pairs are the samples with the same label in the same batch. By exploiting the semantic relations among sample pairs from different domain, the feature differences between different forgery methods are minimized, thus reducing feature shifts when the model is trained based on a limited number of new task samples. $L_{SCL}$ is formally given as:
\begin{equation}
\begin{split}
L_{SCL} =& -\sum_{i=1}^{B} \frac{1}{\sum_{j=1}^B \mathbb{I}_{ \left [i \neq  j \right ] }\mathbb{I}_{ \left [{y}_{i} = {y}_{j} \right ]}} \sum_{j=1}^{B}\mathbb{I}_{ \left [i \neq  j \right ] }\mathbb{I}_{ \left [{y}_{i} = {y}_{j} \right ] } \\
  & \cdot log\left [ \frac{exp(s_{i,j}/\tau)}{exp(s_{i,j}/\tau)+\sum_{k=1}^{B}\mathbb{I}_{ \left [{y}_{i} \neq {y}_{k} \right ] } exp(s_{i,k}/\tau) }  \right ] 
\label{LSCL}
\end{split}
\end{equation} 
where $B$ is the batch size; $y_i$, $y_j$, and $y_k$ are the label of sample $x_i$, sample $x_j$ and sample $x_k$ respectively. $\mathbb{I}$ is the Indicator function, which equals to 1 when the condition in the parentheses holds, otherwise it equals to 0. Thus, $\mathbb{I}_{ \left [i \neq  j \right ] }\mathbb{I}_{ \left [{y}_{i} = {y}_{j} \right ]}$ selects all the positive samples for anchor $x_i$ in the same batch, and $\mathbb{I}_{ \left [{y}_{i} \neq {y}_{k} \right ] }$ selects the negative samples for anchor $x_i$. Score $s_{i,j} = \frac{r_{x_i}^T \cdot r_{x_j}}{\left \| r_{x_i} \right \| \left \| r_{x_j} \right \|} $ is the cosine similarity between the $r_{x_i}$ and $r_{x_j}$. Recall that $r_{x_i}$ and $r_{x_j}$ are the feature representation output by the encoder for sample $x_i$ and sample $x_j$, respectively. $\tau$ is a temperature parameter.



\subsection{Preserving by Multi-Perspective Knowledge Distillation}
\label{sec:preserving}
To further mitigate the catastrophic forgetting issue, we adopt the \textbf{multi-perspective knowledge distillation} to guide the student model to mimic the teacher model, thus maintaining the model's performance on old tasks.

\textbf{Soft Label Knowledge Distillation} Soft label knowledge distillation loss ($KD$) can be seen as a modified cross-entropy loss that considers probabilities over all classes. The purpose of $KD$ loss is to make the output probability of the student model close to the output of the teacher's network. The $KD$ loss is computed as:
\begin{equation}
 L_{KD} = -\sum_{i=1}^B\sum_{j=1}^{C}{ p_{x_i(j)}^t\cdot logp(p_{x_i(j)}^s)}
 \label{kd}
 \end{equation}
where $C$ is the number of classes, $p_{x_i}^t$ and $p_{x_i}^s$ represent the output of the teacher model and the student model, respectively, parameterized by a temperature $T$ as,
 \begin{equation}
p_{x_i(j)}^t = \frac{exp(z_{x_i(j)}^t/T)}{\sum_{k}exp(z_{x_i(k)}^t/T) } 
 \end{equation}
 \begin{equation}
p_{x_i(j)}^s = \frac{exp(z_{x_i(j)}/T)}{\sum_{k}exp(z_{x_i(k)}/T) } 
 \end{equation}
where $z_{x_i}^t$ and $z_{x_i}$ represent the logits output by the teacher model and the student model, respectively. Please note that $p_{x_i}^s$ is slightly different from $p_{x_i}$ in Eq~\ref{eq:LCE}. $p_{x_i}$ is computed by setting $T=1$ to serve as the final prediction for deepfakes detection, while $p_{x_i}^s$ can be computed with varying $T$ (\emph{e.g.}, 20) for model regularization. More results can be found in supplementary materials. Here we use superscript $s$ to distinguish from the prediction output $p_{x_i}$ of the student model.

\textbf{Feature Knowledge Distillation} As complementary to the soft labels generated by the classifier $g(\cdot)$, we further have the student model to learn the intermediate features output by the encoder $f(\cdot)$ of the teacher model. The intuition behind this is that the intermediate feature representations contain more information than the final soft probability output. As pointed out by previous work~\cite{zhao2021multi}, the forgery traces will be present in all levels of the model, which motivates us to use the intermediate features generated by the encoder $f(\cdot)$ as an additional regularization to preserve the previous knowledge. Formally, the $FD$ loss is calculated as :
 \begin{equation}
L_{FD} = \sum_{i=1}^B {\left \| r_{x_i}^t-r_{x_i}^s \right \|_2^2 } 
\label{fd}
 \end{equation}
where $r_{x_i}^t$ denotes the feature representation output of the teacher model and $r_{x_i}^s$ denotes the feature representation output of the student model.


\subsection{Replay Strategy}

\label{sec:replay}
A replay set, denoted by $R$, is usually composed of representative past samples, the selection of which has a significant impact on incremental learning algorithms. Unlike previous methods where $R$ is only used to prevent forgetting past knowledge by replaying past samples, our method further aims to learn domain-generalized representations by aligning $D_i$ with $R$. Existing replay selection strategies are struggling to satisfy our requirements, we thus present a new approach by selecting past samples with high information entropy. Information entropy is a measure of how much information content is contained in a variable \cite{2022Few}. The information entropy of a sample $x_i$ can be computed as:
\begin{equation}
H(x_i) = -\sum_{j=1}^{C} p_{x_i(j)} \cdot log(p_{x_i(j)})
 \end{equation}
where $p_{x_i(j)}$ is the probability of $x_i$ belonging to class $j$, and $C$ is the number of classes. We estimate $p_{x_i(j)}$ by passing $x_i$ to the teacher model so that samples with low prediction confidence will have a high information entropy score. These samples usually cannot be accurately identified by the model \cite{2022Few}, so are considered as informative negative samples that are crucial for contrastive learning \cite{khosla2020supervised}. As a complement, we also select representative past samples that are closest to the real and fake centroids. To be more specific, we select the top $K/4$ from each group, \emph{i.e.}, hard real/fake samples based on information entropy and central real/fake samples based on distance to class centroids, forming a replay set of $K$ samples.

\subsection{DFIL on Task Sequence \emph{D\textsubscript{C}}}
\label{sec:algorithm}
We have introduced our proposed incremental learning for a new task $D_i$. Here we describe the overall process on a sequence of deepfake tasks $D_C$. As illustrated in Algorithm~\ref{alg:algorithm}, $Model^{(1)}_{IL}$ will only be trained using $L_{CE}$ and $L_{SCL}$ as $D_1$ is the first task and there is no previous knowledge to be preserved. 
For each subsequent $Model^{(i)}_{IL}$, we select $K$ samples from $D_{i-1}$ based on $Model^{(i-1)}_{IL}$ as described in Section~\ref{sec:replay}, to update the replay set.
Thereafter, we update $Model^{(i-1)}_{IL} \rightarrow Model^{(i)}_{IL}$ by incrementally training on $R\bigcup D_i$ with our proposed objectives, $L_{CE}$, $L_{SCL}$, $L_{KD}$, and $L_{FD}$. We will show that our proposed $Model^{(i)}_{IL}$ can effectively detect forged samples in $D_i$ as well as in past tasks.


\begin{algorithm}[t]
    \caption{Deepfake incremental learning on \emph{D\textsubscript{C}}. }
    \label{alg:4}
    \begin{algorithmic}[1]
        \REQUIRE A deepfake detection model consisting of a feature extractor $f_\theta$ parameterized by $\theta$ and a linear classifier $g_\phi$ parameterized by $\phi$. A sequence of deepfake tasks $D_C = \left \{D_1,D_2,..., D_N  \right \}$. Batch size $B$. Learning rate $\eta$. Balancing factors $\alpha$, $\beta$, and $\gamma$.
        \STATE Initialize network $Model^{(1)}_{IL} = f_{\theta_1} \circ g_{\phi_1}$. 
        \STATE Set replay set to empty: $R= \emptyset $. 

        \FOR{ $i = 1,2,3,...,N$}
            \STATE Construct dataset $D_i^\prime \Leftarrow D_i \cup R$
            \FOR{$e = 1,2,3,..., |D_i^\prime|/B$}
                \STATE Draw a mini-batch $Batch_{e} = \left \{ (x_j,y_j) \right \} _{j=1}^{B}$ from $D_i^\prime$.
                \STATE \textcolor{blue}{/* learning domain-invariant representations */}\;
                \STATE $ L_{CE} \Leftarrow L_{CE} (Batch_{e}, \theta_i^{e - 1}, \phi_i^{e - 1}) $ (Eq. \textcolor{blue}{\ref{eq:LCE}})
                \STATE $L_{SCL} \Leftarrow L_{SCL} (Batch_{e},\theta_i^{e - 1})$ (Eq. \textcolor{blue}{\ref{LSCL}}) 
                \STATE Compute $L$ by $L  \Leftarrow L_{CE} + \alpha L_{SCL}$ 
                \IF{$i > 1$}
                    \STATE \textcolor{blue}{/* preserving past model's knowledge */}\;
                    \STATE  $L_{KD} \Leftarrow L_{KD} (Batch_e,\theta_{i-1},\theta_{i}^{e - 1},\phi_{i-1},\phi_{i}^{e - 1})$ (Eq. \textcolor{blue}{\ref{kd}})
                    
                    \STATE  $L_{FD} \Leftarrow L_{FD}(Batch_e,\theta_{i-1},\theta_{i}^{e - 1} )$ (Eq. \textcolor{blue}{\ref{fd}})
                    \STATE Update $L$ by $L  \Leftarrow L + \beta L_{KD} + \gamma L_{FD} $ 
                \ENDIF

             
            \STATE Update $\theta_{i}^{e - 1}$ by $\theta_{i}^{e} \Leftarrow \theta_{i}^{e-1} - \eta \nabla_{\theta_{i}^{e-1}}  L$  
            \STATE Update $\phi_{i}^{e - 1}$ by $\phi_{i}^{e} \Leftarrow \phi_{i}^{e-1} - \eta \nabla_{\phi_{i}^{e-1}}  L$ 
            \ENDFOR
            
            \STATE \textcolor{blue}{/* Replay set preparation for next task $D_{i+1}$*/}\;
            \STATE Create $R_i$ by selecting $K/2$ hard samples and $K/2$ center samples from $D_i$
            \STATE Update $R=R\bigcup R_i$
            \STATE $Model^{(i)}_{IL} = f_{\theta_i} \circ g_{\phi_i}$ with $(\theta_i=\theta_i^{|D_i^\prime/B|},\phi_i=\phi_i^{|D_i^\prime/B|})$.

        \ENDFOR

    \end{algorithmic}
    \label{alg:algorithm}
\end{algorithm}

%






\section{Evaluation}
\subsection{Experimental Setup}

\textbf{Dataset} We evaluate our proposed method on four publicly available face deepfake datasets, namely Faceforencis++ (FF++), Celeb-DFv2 (CDF2), Deepfake Detection (DFD) and DeepFake Detection Chanllenge (DFDC). The FF++ dataset is widely used in forgery detection tasks, which contains four forgery methods, namely DeepFakes, Face2Face, FaceSwap, and Neural-Textures. The FF++ dataset includes a total of 4000 fake videos and 1000 real videos. CDF2 contains 590 real videos and 5639 fake videos. We use DFDC-Preview from DFDC, which contains 5214 videos and two forgery methods. 
The task sequence used in our experiment is set to $D_C=\{$FF++, DFDC-P, DFD, CDF2$\}$. 
Table~\ref{tab:dataset} summarizes the statistics of the four datasets. While the whole training set is kept for the first task FF++, we randomly select 25 training fake sample videos from each subsequent task for incremental learning based on limited samples.

\begin{table}
\small
\centering
\caption{Datasets Description. The details of the total number of videos used for training and test.}
\begin{tabular}{c|ccccc} 
\hline
Datasets & \begin{tabular}[c]{@{}c@{}}Total \\Videos\end{tabular} & \begin{tabular}[c]{@{}c@{}}Train \\Videos\end{tabular} & \begin{tabular}[c]{@{}c@{}}Validation \\Videos\end{tabular} & \begin{tabular}[c]{@{}c@{}}Continual\\~Learning\end{tabular} & \begin{tabular}[c]{@{}c@{}}Test\\~Videos\end{tabular}  \\ 
\hline
    FF++     & 5000  & 3000 & 1000 & - & 1000 \\ 
\hline
    DFDC-P   & 5214  & 3149 & 700  &  25 & 1052 \\ 
\hline
    DFD      & 1000  & 700  & 150  & 25  & 150  \\ 
\hline
    CDF2     & 6229  & 3917 & 1307 & 25  & 1305\\
\hline
\end{tabular}
\label{tab:dataset}
\end{table}



\begin{table}
\small
\centering
\caption{Performance comparison of backbones on FF++.}
\begin{tabular}{c|ccc}
\hline
Method             & EfficientNet-B4           & ResNet34 &    \textbf{Xeception}       \\ 
\hline
   ACC  & 95.43          &     93.52    & \textbf{95.52}    \\ 
\hline
   AUC   &0.9818     &        0.9666     & \textbf{0.9855}       \\ 
\hline
\end{tabular}
\label{tab:backbone}
\end{table}

\textbf{Implementation Details}
In the data processing stage, we adopt the advanced face extractor RetinaFace \cite{2020RetinaFace} to obtain face regions and then use the extracted faces for model training and testing. The replay set size K is set to 500. We trained our model for 20 epochs on task 1 and continued training for 20 epochs for each subsequent task. We adopt the Adam optimizer, and set the learning rate to 5e-4, which decays by 0.5 in every five epochs.

\textbf{Evaluation metric}
We adopt four commonly used evaluation metrics described below.

(1) Accuracy (ACC) calculates the ratio between the correctly predicted samples and all samples by $ACC = \frac{TP+TN}{n}$ where $TP$ and $TN$ refer to the true positives and true negatives, respectively. $n$ is the total number of images.

(2) Average Accuracy (AA) refers to the average accuracy of all previous tasks, calculated as $AA = \frac{1}{N} \sum_{i}^{N} ACC_i$ where $ACC_i$ is the ACC of $i$-th task and $N$ is the number of tasks.

(3) Average Forgetting (AF) refers to the average forgetting rate of previous tasks. 
It is calculated as $AF = \frac{1}{N} \sum_{i}^{N} (ACC_i^{first} - ACC_i^{last})$ where $ACC_i^{first}$ refers to the accuracy of task $i$ after training on task $i$ for the first time, $ACC_i^{last}$ refers to the accuracy of task $i$ after training on the current task.

(4) Area Under the ROC Curve (AUC) computes the two-dimensional area underneath the ROC curve. 
We use this measure to evaluate the generalization ability of different forgery detection methods across tasks.

\subsection{Comparison with Existing Incremental Learning Method}

For evaluation, we compare our proposed method against five few-shot incremental learning methods described below. 
\begin{itemize}

\item \textbf{LWF:} Learning without forgetting \cite{li2017learning}, a regularization method that uses knowledge distillation to constrain the feature drifts on previous datasets.

\item \textbf{EWC:} Elastic weight consolidation \cite{kirkpatrick2017overcoming}, an approach that remembers old tasks by selectively slowing down the learning on the weights that are important for those tasks.

\item \textbf{DGR:} Deep Generative Replay \cite{shin2017continual}, a replay generation method that uses old data to train a generative model. 

\item \textbf{ER:} Experience Replay \cite{chaudhry2019tiny}, a replay selection method that randomly selects old task samples into the replay set. 

\item \textbf{SI:} Synaptic Intelligence \cite{zenke2017continual}, a method that brings biological complexity into artificial neural networks. 

\item \textbf{Offline:} A naive implementation that trains the model over the union of all tasks.

\item \textbf{Finetune:} It finetunes a model based on new task data only without addressing the catastrophic forgetting issue. 
\end{itemize}

\begin{table}
\small
\centering
\caption{Performance of Finetune and Offline on few-shot incremental learning. Finetune method suffers from catastrophic forgeting. Offline method provides the upper bound.}
\begin{tabular}{c|c|cccc|cc} 
\hline
Method      & Dataset & FF++  & DFDC-P & DFD   & CDF2  & AA    & AF     \\ 
\hline
 \multirow{4}{*}{FT}    & FF++    & 95.53 & -      & -     & -     & 95.53 & -      \\ 
\cline{2-2}
     & DFDC-P  & 89.40 & 79.14  & -     & -     & 84.27 & 6.49   \\ 
\cline{2-2}
                   & DFD     & 69.16 & 43.58  & 94.69 & -     & 69.14 & 31.15  \\ 
\cline{2-2}
                             & CDF2    & 66.34 & 62.16  & 73.56 & 81.13 & 70.79 & 22.55  \\ 
\hline
OL & All     & 95.39 & 79.19  & 96.26 & 89.55 & 90.09 & -      \\
\hline
\end{tabular}
\label{tab:offline-finetune}
\end{table}

\textbf{Comparison of Backbone networks} 
Existing incremental learning methods were initially proposed for different tasks and implemented with different backbones. For a fair comparison, we conduct experiments using the same backbone model in all methods. Three different models, namely Xception, Resnet34, and EfficientNet, are evaluated. Their performance on the FF++ dataset is reported in Table~\ref{tab:backbone}. 
The results show that all three models can achieve good forgery detection results, among which Xception obtains the highest ACC of 95.52 and AUC of 0.9855. Therefore, Xception is adopted as the backbone model in the rest of our experiments.

\textbf{Performance on deepfakes Incremental learning} We first perform incremental learning using Finetune (FT) and Offline (OL) and report the results in Table~\ref{tab:offline-finetune}. The Finetune method updates the model with cross-entropy loss based on new task data only. Thus, it quickly loses its ability on detecting previous forgery samples after learning on new tasks, obtaining an AF of 22.55 only after training the last task CDF2. Moreover, the model's performance on subsequent tasks is less satisfactory than its performance on the first task. This is because the training samples for DFDC-P, DFD, and CDF2 are quite limited, indicating the importance of our proposed deepfake incremental learning method that can generalize well to new tasks even when only a few samples are available.
The Offline method trains the model with cross-entropy loss using the data of all four tasks. We provide this result to serve as the upper bound for incremental learning tasks.

Next, we compare our proposed method with five existing incremental learning strategies and report the results in Table~\ref{tab:IL}. As it shows, all the existing incremental learning methods are struggling to perform consistently well on the new task. As they focus on learning with new data, on the one hand, they are less effective in preserving the past knowledge learned on the old tasks. On the other hand, they tend to overfit the new task due to the small number of new task samples that are available for training. Comparatively, in terms of preserving past knowledge, our proposed method obtains the best AF of 7.01 after training on all four tasks, outperforming existing methods by a large margin. In terms of learning on new tasks, our method prevents overfitting by aligning the representations of old and new task data based on supervised contrastive learning. We obtain the best AA of 85.49, indicating that our method can recognize the forgery samples in new tasks with better consistencies than existing incremental learning methods.


 \begin{small}
\begin{table}
\centering
\caption{Performance comparison on few-shot deepfake incremental learning.}
\begin{tabular}{c|c|cccccc} 
\hline
Method       & Dataset & FF++  & DFDC-P & DFD   & CDF2  & AA$\uparrow$     & AF$\downarrow $     \\ 
\hline
             & FF++    & 95.52     & -      & -     & -     & 95.52     & -     \\ 
\cline{2-2}
             & DFDC-P  & 87.83     & 81.57      & -     & -     & 84.70     & 7.69       \\ 
\cline{2-2}
LWF          & DFD     & 76.16     & 41.78      & 96.36     & -     & 71.43     & 19.89      \\ 
\cline{2-2}
             & CDF2    & 67.34     & 67.43      & 84.05     & 87.90     & 76.68     & 14.44      \\ 
\hline
             & FF++    & 95.51     & -      & -     & -     & 95.51     & -      \\ 
\cline{2-2}
             & DFDC-P  & 92.74     & 84.85      & -     & -     & 88.79     & 2.77      \\ 
\cline{2-2}
EWC          & DFD     & 90.76     & 48.18      & 95.21     & -     & 78.05     & 20.70      \\ 
\cline{2-2}
             & CDF2    & 87.88     & 67.53      & 85.94     & 81.89     & 80.81     & 11.04      \\ 
\hline
             & FF++    & 88.86     & -      & -     & -     & 88.66     & -       \\ 
\cline{2-2}
             & DFDC-P  & 78.81     & 83.89      & -     & -     & 81.35     & 10.05      \\ 
\cline{2-2}
DGR          & DFD     & 64.31     & 73.31      & 89.69     & -     & 75.57     & 17.56      \\ 
\cline{2-2}
             & CDF2    & 67.33     & 79.65      & 78.35     & 76.50     & 75.45     & 12.37      \\ 
\hline
             & FF++    & 95.52     & -      & -     & -     & 95.52     & -      \\ 
\cline{2-2}
             & DFDC-P  & 92.25     & 88.53      & -     & -     & 90.39     & 3.27       \\ 
\cline{2-2}
ER          & DFD     & 84.69     & 80.59      & 94.00     & -     & 86.42     & 9.38      \\ 
\cline{2-2}
             & CDF2    & 71.65     & 69.40      & 92.98     & 83.26     & 79.32     & 14.67      \\ 
\hline
             & FF++    & 95.51     & -      & -     & -     & 95.51     & -      \\ 
\cline{2-2}
             & DFDC-P  & 90.88     & 88.38      & -     & -     & 89.63     &  4.63      \\ 
\cline{2-2}
SI           & DFD     & 45.46     & 27.02      & 96.67     & -     & 56.38     & 55.70      \\ 
\cline{2-2}
             & CDF2    & 36.67     & 42.04      & 44.83     & 78.88     & 50.60     & 50.79      \\ 
\hline
             & FF++    & 95.67 & -      & -     & -     & \textbf{95.67} & -      \\ 
\cline{2-2}
             & DFDC-P  & 93.15 & 88.87  & -     & -     & \textbf{91.01} & \textbf{2.52}      \\ 
\cline{2-2}
\textbf{Ours} & DFD     & 90.30 & 85.42  & 94.67 & -     & \textbf{90.03} & \textbf{4.41}      \\ 
\cline{2-2}
             & CDF2    & 86.28 & 79.53  & 92.36 & 83.81 & \textbf{85.49} & \textbf{7.01}      \\
\hline
\end{tabular}
\label{tab:IL}
\end{table}
\end{small}

\subsection{Ablation Study}

\textbf{Effect of $L_{SCL}$, $L_{KD}$ and $L_{FD}$} 
We study the effect of different losses in our method and report the results in Table~\ref{tab:ablation_losses}. First, we let the student model learn the feature representations of the teacher model through Loss $L_{FD}$ (FT+FD). The results show that regularizing the student model based on the teacher model's intermediate features only is not sufficient to prevent catastrophic forgetting. Next, we use multi-perspective knowledge distillation (FT+FD+KD), which slows down the degree of forgetting compared to using $L_{FD}$ or $F_{KD}$ only. Then we add a replay mechanism to further stop catastrophic forgetting, the effect of which will be discussed in the next paragraph. Finally, we introduce our proposed supervised contrastive learning objective $L_{SCL}$ to learn domain-invariant representations (Ours refers to FT+FD+KD+SCL+Replay). We can see that our proposed method achieves the best detection results, which verifies the effectiveness of each objective function we used in our method.

\textbf{Effect of Replay Set} 
As shown in Table~\ref{tab:ablation_losses}, we study the effect of the replay set by comparing the results obtained with and without it in our method. Without the replay set (Ours w/o R), the supervised contrastive learning is performed based on new task samples only. The generalization ability of the learned representations is unsatisfactory, resulting in catastrophic forgetting on the old tasks.


\begin{table}
\small
\centering
\caption{Ablation study on the effect of our proposed objectives. The best performer is highlighted in boldface.}

\begin{tabular}{c|c|cc|cc|cc} 
\hline
\multirow{2}{*}{Method} & FF++ & \multicolumn{2}{c|}{DFDC-P}     & \multicolumn{2}{c|}{DFD}     & \multicolumn{2}{c}{CDF2}      \\ 
\cline{2-8}
                        & AA                    & AA             & AF            & AA             & AF            & AA             & AF             \\ 
\hline
FT                      & 95.53                & 84.27          & 6.49          & 69.14          & 31.15         & 70.79          & 22.55          \\ 
\hline
FT+FD                   & 95.53                 & 86.83          & 10.81         & 72.72          & 29.56         & 62.55          & 38.15          \\ 
\hline
FT+FD+KD                & 95.53               & 89.17          & 5.98          & 75.45          & 25.80         & 64.36          & 35.20          \\ 
\hline
\begin{tabular}[c]{@{}c@{}}FT+FD+\\KD+Replay\end{tabular}         & 95.52                   & 87.85          & 5.52          & 87.20          & 7.39          & 85.37          & 7.68           \\ 
\hline
Ours w/o R                & 95.67               & 90.64          &3.38          & 70.26          & 34.61         & 68.37          & 30.27          \\ 
\hline
\textbf{\textbf{Ours}}   & \textbf{95.67}          & \textbf{91.01} & \textbf{2.52} & \textbf{90.03} & \textbf{4.41} & \textbf{85.49} & \textbf{7.01}  \\
\hline
\end{tabular}
\label{tab:ablation_losses}
\end{table}


\begin{table}
\small
\centering
\caption{Performance comparison of different replay strategies. The best result is highlighted in boldface.}
\begin{tabular}{c|c|cc|cc|cc} 
\hline
\multirow{2}{*}{\begin{tabular}[c]{@{}c@{}}Replay\\Strategy\end{tabular}} & FF++ & \multicolumn{2}{c|}{DFDC-P} & \multicolumn{2}{c|}{DFD} & \multicolumn{2}{c}{CDF2}  \\ 
\cline{2-8}
 & AA                    & AA & AF                    & AA & AF                    & AA & AF                     \\ 
\hline
All Hard~ ~                                                             & 95.67                     & 90.54  & \textbf{1.34}                     & 89.63  & \textbf{4.10}                     & 83.23  & 9.36                      \\ 
\hline
All Easy                                                                & 95.67                    & 89.77  & 4.61                     & 86.18  & 10.46                     & 81.84  & 12.47                      \\ 
\hline
All Margin                                                             & 95.67                       & 90.68  & 2.55                     & 82.94  & 15.07                     & 79.12  & 14.59                      \\ 
\hline
All Center                                                              & 95.67                   & 90.98  & 2.27                     & 85.66  & 11.16                     & \textbf{86.04}  & 7.51                      \\ 
\hline
\textbf{\textbf{Ours}}                                                   & \textbf{95.67}                  & \textbf{91.01}  & 2.52                     & \textbf{90.03}  & 4.41                     & 85.49  & \textbf{7.01}                      \\
\hline
\end{tabular}
\label{tab:replay}
\end{table}

\begin{figure*}[t] 
\centering 
\includegraphics [width=0.9\textwidth]{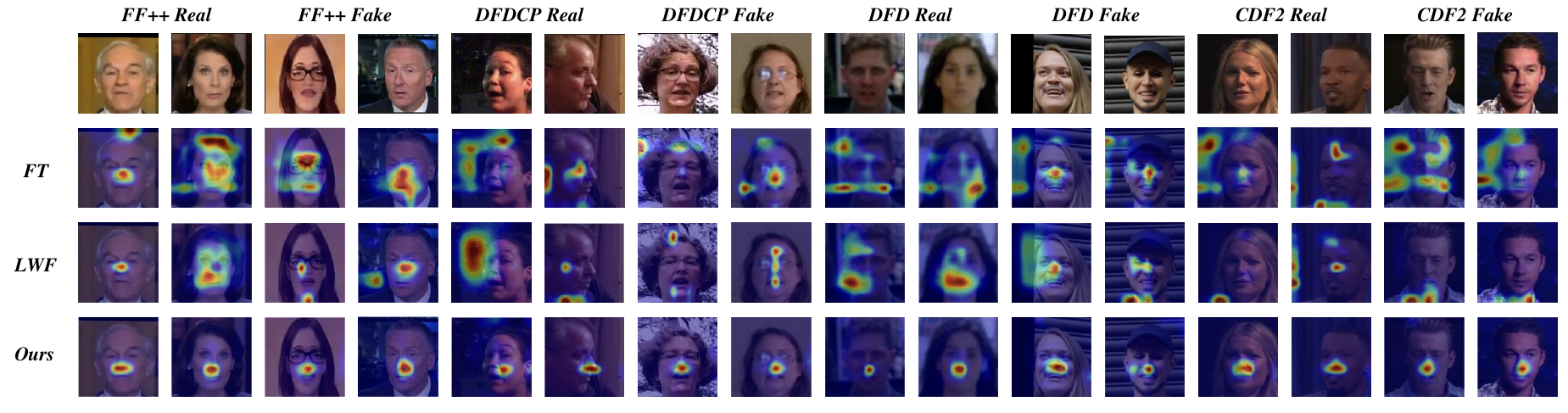} 
\caption{Comparison of feature visualization on different datasets. The first row contains real and fake images from different datasets.The remaining three rows show the visualization results of the three methods. } 
\label{fig:visualize} 
\end{figure*}

\textbf{Effect of Different Replay Strategies}
We investigate different strategies to select the replay samples. The first strategy for selecting replay samples is to select all real and fake samples with high information entropy (All Hard). Specifically, for real samples, we select the top $K/2$ samples with the highest information entropy, and for fake samples, we also select the top $K/2$ samples with the highest information entropy. The second selection strategy is to select all samples with the lowest information entropy (All Easy). Similarly, we select the top $K/2$ samples with the lowest information entropy from real and fake samples, respectively. In the third selection strategy, we select the samples that are farthest from the class centroids (All Margin). 
$K/2$ real (and fake) samples are selected that are farthest from the centroid of the real (and fake) class. In the fourth selection strategy, we select the samples that are closest to the class centroids (All Center). $K/2$ real (and fake) samples are selected that is closest to the centroid of the real (and fake) class. The results obtained by different selection strategies are presented in Table~\ref{tab:replay}. 
The All Hard method has higher AF indicators in DFDC-P and DFD, but lower AA indicators than ours. This is because All Hard method selects a large number of difficult samples, resulting in the model paying too much attention to the previous tasks. Thus, the model does not generalize well to new samples, resulting in low average accuracy. Moreover, because of the relatively low accuracy, the magnitude of forgetting on past tasks tends to be less significant when continually learning new tasks. This can be another reason why All Hard method achieves a good average forgetting rate in DFDC-P and DFD. For All Center method, it obtains a higher average accuracy in CDF2 than ours, but performs less effectively than ours in terms of the average forgetting rate. This is because the central samples only are not representative enough for the old task, resulting in the model forgetting the previous knowledge more quickly. Finally, All Easy and All Margin perform less effectively than ours in terms of both AA and AF. 


\subsection{Visualization}
We employ LayerCAM \cite{jiang2021layercam} to visualize the feature representations learned by our method, the FT method, and the LWF method. The visualized models underwent three deepfakes incremental learning tasks. 
Figure~\ref{fig:visualize} visualizes four examples in each data where our method successfully detects their correct label, while the other two methods fail. 
We can see that the features generated by FT and LWF are less consistent, sometimes failing to pay attention to the facial area, resulting in incorrect predictions. Comparatively, our method aligns the features across tasks based on $L_{SCL}$, focusing better on the facial area and the forged traces.


\subsection{Generalization Comparison}
Finally, we compare the generalization ability of our method with existing deepfakes detection models and use AUC as the evaluation metric. The results are reported in Table~\ref{tab:generalization}. As can be seen, existing deepfakes detection models are trained on dataset FF++, thus their performance on subsequent tasks degrades significantly due to domain shifts. Comparatively, by leveraging 25 sample videos from each of the subsequent tasks, our method improves the generalization ability of deepfake detection model significantly, obtaining a much better AUC on all four datasets. After learning on all the four tasks, our model still retains excellent detection performance on the second task DFDCP, obtaining an AUC of 0.9173 with an absolute improvement of 0.1203 compared to the second best method~\cite{luo2021generalizing}.
On the third task DFD, our method's AUC is 0.9756, which is 0.288 higher than the second best method~\cite{zhuang2022uia}. On the last task CDF2, our method obtains an AUC of 0.8968, outperforming the second best method~\cite{dong2022protecting} by 0.397. The experimental results indicate that our proposed incremental learning method can significantly improve the generalization ability of deepfake detection model facilitated with a small number of new task samples.

\begin{table}
\small
\centering
\caption{Comparison of generalization ability of different deepfakes detection methods.}
\begin{tabular}{clccc} 
\hline
{Method}             & ~ ~~~~ ~ ~ ~~ & DFDCP          & DFD            & CDF2             \\ 
\hline
Xception \cite{rossler2019faceforensics++}        &            & 0.7220          & 0.7050          & 0.6550           \\ 
\hline
Face X-ray \cite{li2020face}      &            & 0.7000          & 0.9347          & 0.7520           \\ 
\hline
Luo.et.al. \cite{luo2021generalizing}      &            & 0.7970          & 0.9190          & 0.7940           \\ 
\hline
Multi Attention \cite{zhao2021multi} &            & 0.6628          & 0.7553          & 0.6744           \\ 
\hline
LTW \cite{sun2021domain}             &            & 0.7458          & 0.8856          & 0.7717           \\ 
\hline
Local-realtion \cite{chen2021local}  &            & 0.7653          & 0.8924          & 0.7826           \\ 
\hline
DCL \cite{sun2022dual}            &            & -              & 0.9166          & 0.8230           \\ 
\hline
ICT \cite{dong2022protecting}            &            & -              & 0.8413          & 0.8571           \\ 
\hline
SOLA \cite{fei2022learning}           &            & -              & -              & 0.7205           \\ 
\hline
UIA-ViT \cite{zhuang2022uia}         &            & 0.7580          & 0.9468          & 0.8241           \\ 
\hline
\textbf{Ours}    &            & \textbf{0.9173} & \textbf{0.9756} & \textbf{0.8968}  \\
\hline
\end{tabular}
\label{tab:generalization}
\end{table}

\section{Conclusion}
We propose an new incremental learning method for deepfakes detection, which leverages supervised contrastive learning to obtain domain-generalized representations among the old and new task samples. When learning a new task, a carefully designed replay set is utilized to reduce the feature distance between the samples that belong to the same class across domains. Through extensive experimental evaluation, we show that our method is more effective than existing incremental learning approaches when dealing with a limited number of new task samples. Moreover, our method can effectively improve the generalization ability of deepfakes detection models, which is critical for forgery detection applications in real-world scenarios.

\begin{acks}
This work is partially supported by the National Key R\&D Program of China (2020AAA0107700), the National Natural Science Foundation of China (62172359) and the Key R\&D Program of Zhejiang Province (No. 2023C01217).
\end{acks}
\balance
\bibliographystyle{plain}

\bibliography{myref.bib}

\newpage

\appendix

\section{Preliminaries and Object Function}
In EWC \cite{kirkpatrick2017overcoming}, authors use the Bayesian formula as the incremental learning objective function, which can be written as: 

\begin{equation}
\begin{split}
   logp(\theta |D_{t-1},D_t) & = log\frac{p(\theta D_{t-1}D_{t})}{p(D_{t-1}D_{t})}  \\
                            & =  log\frac{p(\theta D_{t-1}|D_{t}) p(D_{t})}{p(D_{t-1}|D_{t})p(D_{t})}  \\
                            & =  log\frac{p(\theta D_{t-1}|D_{t})}{p(D_{t-1}|D_{t})} \\
                            & = log\frac{p(D_{t}| \theta D_{t-1})p(\theta| D_{t-1})}{p(D_{t-1}|D_{t})} \\
                            & =  logp(D_t|\theta )+logp(\theta |D_{t-1})+logp(D_t)
   \label{ewc_1}
\end{split}
\end{equation}
where $D_{t-1}$ and $D_t$ denote task $t-1$ data and task $t$ data, respectively. $\theta$ is the parameters of the model, $logp(D_t|\theta)$ is the likelihood of task $D_t$, $logp(D_t)$ is a constant, $logp(\theta|D_{t-1})$ is the posterior of task $t$. If we want to maximize $logp(\theta |D_{t-1},D_t)$, it is equivalent to maximize the last three, and because $logp(D_t)$ is a constant, Eq. \ref{ewc_1} can be rewritten as :
\begin{equation}
   \underset{\theta}{argmax~} logp (\theta |D_{t-1},D_t) 
   = \underset{\theta}{argmax~} (log p(D_t|\theta ) + logp(\theta |D_{t-1})) \label{ewc}
\end{equation}
So the goal of incremental learning is to optimize $logp(\theta|D_{t-1})$ and $logp(D_t|\theta)$. 
Optimizing $logp(D_t|\theta)$ is to learn the new task maximally, while optimizing $logp(\theta|D_{t-1})$ is to minimize the degree of forgetting the previous tasks.

The overall loss function of DFIL is given as
 \begin{equation}
L_{DFIL} = \underbrace{L_{CE}+ \alpha L_{SCL}}_{(1)Learning}  + \underbrace{\beta L_{KD} + \gamma L_{FD}}_{(2)Preserving} 
 \end{equation}

The goal of optimizing $logp(D_{t}|\theta)$ in Eq. \ref{ewc} is actually to minimize the loss function of the model on task $t$, that is, to minimize $L_{CE}$ and $L_{SCL}$, so minimizing $L_{CE}$ and $L_{SCL}$ is equivalent to maximizing $logp(D_{t}|\theta)$.

In Eq. \ref{ewc}, if we want to optimize $log(\theta|D_{t-1})$, it means that we need to preserve the performance of the model for dataset $D_{t-1}$. We use $L_{KD}$ and $L_{FD}$ loss functions to constrain the student model to learn the performance of the old task dataset $D_{t-1}$ by learning the output content (soft labels and feature representations) of the teacher model. So it is equivalent to optimize $log(\theta|D_{t-1})$.

\section{Additional Experiments}
 \begin{small}
\begin{table}
\centering
\caption{More detail performance comparison of different replay strategies. The best result is highlighted in boldface.}
\begin{tabular}{c|c|cccccc} 
\hline
Method       & Dataset & FF++  & DFDC-P & DFD   & CDF2  & AA$\uparrow$     & AF$\downarrow $     \\ 
\hline
             & FF++    & 95.67     & -      & -     & -     & 95.52     & -     \\ 
\cline{2-2}
All              & DFDC-P  & 94.33     & 86.76      & -     & -     & 90.54     & \textbf{1.34}       \\ 
\cline{2-2}
Hard          & DFD     & 91.76     & 82.48      & 94.67     & -     & 89.63     & \textbf{4.10}      \\ 
\cline{2-2}
             & CDF2    & 82.31     & 78.12      & 88.57     & 83.95     & 83.23     & 9.36      \\ 
\hline
             & FF++    & 95.67     & -      & -     & -     & 95.51     & -      \\ 
\cline{2-2}
All             & DFDC-P  & 91.06     & 88.48      & -     & -     & 88.79     & 4.61      \\ 
\cline{2-2}
Easy          & DFD     & 81.40     & 81.84      & 95.31     & -     & 78.05     & 10.46      \\ 
\cline{2-2}
             & CDF2    & 75.99     & 81.17      & 85.90     & 84.32     & 81.84     & 12.47      \\ 
\hline
             & FF++    & 95.67     & -      & -     & -     & 95.67     & -       \\ 
\cline{2-2}
All            & DFDC-P  & 93.12     & 88.25      & -     & -     & 90.68     & 2.55      \\ 
\cline{2-2}
Margin          & DFD     & 81.30     & 72.47      & 95.06     & -     & 82.94     & 15.07      \\ 
\cline{2-2}
             & CDF2    & 79.97     & 68.34      & 86.16     & 82.03     & 79.12     & 14.59      \\ 
\hline
             & FF++    & 95.67     & -      & -     & -     & 95.67     & -      \\ 
\cline{2-2}
All             & DFDC-P  & 92.97     & 89.00      & -     & -     & 90.98     & 2.27       \\ 
\cline{2-2}
Center           & DFD     & 90.44     & 71.90      & 94.64     & -     & 85.66     & 11.16      \\ 
\cline{2-2}
             & CDF2    & 89.28     & 78.48      & 91.02     & 85.38     & \textbf{86.04}     & 7.51      \\ 
\hline
             & FF++    & 95.67 & -      & -     & -     & \textbf{95.67} & -      \\ 
\cline{2-2}
             & DFDC-P  & 93.15 & 88.87  & -     & -     & \textbf{91.01} & 2.52      \\ 
\cline{2-2}
\textbf{Ours} & DFD     & 90.30 & 85.42  & 94.67 & -     & \textbf{90.03} & 4.41      \\ 
\cline{2-2}
             & CDF2    & 86.28 & 79.53  & 92.36 & 83.81 & 85.49 & \textbf{7.01}      \\
\hline
\end{tabular}
\label{tab:select}
\end{table}
\end{small}

\begin{table}
\centering
\caption{Performance of training a new model by using few data. The best result is highlighted in boldface.}
\begin{tabular}{c|cccc} 
\hline
Method             & FF++ & DFDC-P & DFD   & CDF2   \\ 
\hline
Training A New Model & -    & 82.44 & 95.05 & 74.63  \\
Ours               & -    & \textbf{88.87} & \textbf{96.92}     & \textbf{84.68}      \\
\hline
\end{tabular}
\label{tab:new}
\end{table}

\subsection{More detail performance comparison of different replay strategies}
In Table \ref{tab:select}, we show in more detail the effect of various methods of selecting samples on DFIL. In the All Hard method, the AF of the model after learning the DFDC-P dataset is 1.34, but the ACC of the model on the new task DFDC-P is only 86.76. This is because the model does not finish learning the new task with a small amount of samples. In addition, after model learning the DFD dataset, the model's ACC on DFDC-P drops to 82.48, which is lower than our method (our method's ACC on DFD-P is 85.42). More importantly, after performing the task learning of CDF2, the ACC of the model drops severely on FF++, DFDC-P, and DFD datasets, and the AF of the model reaches 9.36 (our method has only 7.01). Therefore our strategy is to be better than the strategy of selecting all difficult samples (samples with high information entropy). In the All Easy method, although model has an ACC of 95.31 on DFD after learning the DFD dataset, the AF is as high as 10.46. This is because easy samples (samples with low information entropy) have a low impact on the model and are not effective in preventing catastrophic forgetting. In All Margin method, the model also underwent catastrophic forgetting after learning a new task. And in the All Center method, the model's ACC in DFDC-P after learning the DFD task is 71.90 (our method's ACC is 85.42, a decrease of 17.10.) Then the model's AA after learning CDF2 is 86.04, but the AF was 7.51 (7.01 in our method). It indicates that the model with the All Center method is more prone to forgetting newly learned tasks and is not suitable for use with fewer samples. In summary, our method of selecting samples is more suitable for solving the deepfake incremental learning task in the case of a few samples.

\subsection{Comparison of training a new model by using few data.}

In Table \ref{tab:new}, we compare the performance of training a completely new detection model directly using a small amount of data with our method DFIL. In the first row, we initialize a detection model and then train it using a small number of samples from DFDC-P, DFD, and CDF2, respectively. And in the second line, we take DFDC-P, DFD, and CDF2 as the second task and use DFIL for learning, respectively. Our method achieves an ACC of 88.87 for DFDC-P, which is 6.34 higher than that of the retrained detection model. On the DFD task, our method achieves an ACC of 96.92, which is 1.57 higher than that of the retrained detection model, while on the CDF2 task, our method performs even better with an ACC of 84.68, which is 10.05 higher than that of the retrained detection model. In the deepfake incremental learning task with few samples, the detection model is difficult to learn new forgery features with a small amount of data since the new forgery algorithm does not have enough samples to represent whole feature distribution. Meanwhile, If the model is trained directly on top of the old one, catastrophic forgetting can occur. Our proposed method DFIL not only learns new tasks with a small number of samples but also solves the problem of catastrophic forgetting.

\subsection{Different temperature $\tau $ in knowledge distillation}
In Table \ref{tab:t}, we show the effect of different temperature $\tau $ on DFIL. In soft-labeled knowledge distillation, the logits of the teacher model and the student model are divided by the temperature $\tau $, which is then input to the softmax layer to determine the probability, so the temperature $\tau $ affects the student model to learn the old knowledge. From the Table \ref{tab:t} we find that when the temperature $\tau $ is equal to 10, the AA of the model is 83.31 and the AF is up to 8.05 (to be lower than the optimal $\tau = 20$ ) after the model learning the CDF2, despite the fact that the student model learns some old knowledge. When the temperature $\tau $ is 30, the AF of the model reaches 8.16 after model learning the CDF2 dataset (compared to 7.01 for a temperature $\tau = 20 $ ), although the AF of the model is 2.36 and 3.86 for the first two tasks (which lower than the case of a temperature $\tau = 20 $ ). In summary, the analysis of the temperature $\tau = 20$  is the most suitable for the student model for the learning of old knowledge.

\begin{small}
\begin{table}
\caption{Performance comparison of temperature $\tau $.}
\centering
\begin{tabular}{c|c|cccccc} 
\hline
Method          & Dataset & FF++ & DFDC-P & DFD & CDF2 & AA & AF  \\ 
\hline
                & FF++    & 95.67    & -     & -   & -    & 95.67  & -   \\ 
\cline{2-2}
$\tau=10$            & DFDC-P   & 92.95    & \textbf{89.53}     & -   & -    & \textbf{91.24}  & 2.72   \\ 
\cline{2-2}
                & DFD     & 91.11    & 82.78     & 94.40   & -    & 89.43  & 5.65   \\ 
\cline{2-2}
                & CDF2    & 87.00    & 77.66     & 90.78   & 77.82    & 83.31  & 8.05   \\ 
\hline
                & FF++    & \textbf{95.67}    & -     & -   & -    & \textbf{95.67}  & -   \\ 
\cline{2-2}
$\tau=20$            & DFDC-P   & 93.15    & \textbf{88.87}     & -   & -    & 91.01  & 2.52   \\ 
\cline{2-2}
\textbf{(Ours)} & DFD     & 90.30    & \textbf{85.42}     & 94.67   & -    & 90.03  & 4.41   \\ 
\cline{2-2}
                & CDF2    & 86.28    & \textbf{79.53}     & \textbf{92.36}   & \textbf{83.31}    & \textbf{85.49}  & \textbf{7.01}   \\ 
\hline
                & FF++    & 95.67    & -     & -   & -    & 95.67  & -   \\
$\tau=30$            & DFDC-P   & \textbf{93.31}    & 87.03     & -   & -    & 90.17  & \textbf{2.36}   \\
                & DFD     & \textbf{91.62}    & 83.36     & \textbf{96.12}   & -    & \textbf{90.36}  & \textbf{3.86}   \\
                & CDF2    & \textbf{86.35}    & 79.19     & 88.78   & 79.93    & 83.56  & 8.16   \\
\hline
\end{tabular}
\label{tab:t}
\end{table}
\end{small}

\subsection{More detail Ablation study}
In Table \ref{tab:Ab}, we show results of the ablation experiments in more detail. The detailed experimental results further support our conclusions in 5.3 in the main text.

 \begin{small}
\begin{table}
\centering
\caption{More detail results of the ablation experiments. The best result is highlighted in boldface.}
\begin{tabular}{c|c|cccccc} 
\hline
Method       & Dataset & FF++  & DFDC-P & DFD   & CDF2  & AA$\uparrow$     & AF$\downarrow $     \\ 
\hline
             & FF++    & 95.53     & -      & -     & -     & 95.53     & -     \\ 
\cline{2-2}
             & DFDC-P  & 89.40     & 79.14      & -     & -     & 84.27     & 6.69       \\ 
\cline{2-2}
FT          & DFD     & 69.16     & 43.58      & 94.69     & -     & 69.14     & 31.15      \\ 
\cline{2-2}
             & CDF2    & 66.34     & 62.16      & 73.56     & 81.13     & 70.79     & 22.55      \\ 
\hline
             & FF++    & 95.53     & -      & -     & -     & 95.53     & -      \\ 
\cline{2-2}
FT             & DFDC-P  & 84.72     & 88.94      & -     & -     & 86.83     & 10.81      \\ 
\cline{2-2}
+FD          & DFD     & 65.47     & 59.87      & 92.82     & -     & 72.72     & 29.56      \\ 
\cline{2-2}
             & CDF2    & 44.93     & 52.44      & 65.47     & 87.37     & 62.55     & 38.15      \\ 
\hline
             & FF++    & 95.53     & -      & -     & -     & 95.52     & -       \\ 
\cline{2-2}
 FT+FD             & DFDC-P  & 89.55     & 88.79      & -     & -     & 89.17     & 5.98      \\ 
\cline{2-2}
+KD         & DFD     & 73.02     & 59.69      & 93.66     & -     & 75.45     & 25.80      \\ 
\cline{2-2}
             & CDF2    & 45.68     & 62.74      & 63.94     & 85.07     & 64.36     & 35.20      \\ 
\hline
             & FF++    & 95.52     & -      & -     & -     & 95.52     & -      \\ 
\cline{2-2}
 FT+FD            & DFDC-P  & 90.00     & 85.70      & -     & -     & 87.85     & 5.52       \\ 
\cline{2-2}
+KD          & DFD     & 84.45    & 81.99      & 95.18     & -     & 87.20     & 7.39      \\
\cline{2-2}
+Replay             & CDF2    & 78.51    & 82.37      & 92.48     & 88.12     & 85.37     & 7.68      \\ 
\hline
             & FF++    & 95.67     & -      & -     & -     & 95.67     & -      \\ 
\cline{2-2}
Ours             & DFDC-P  & 92.29     & 88.99      & -     & -     & 90.64     &  3.38      \\ 
\cline{2-2}
w/o R           & DFD     & 73.02     & 42.42      & 95.35     & -     & 70.26     & 34.61      \\ 
\cline{2-2}
             & CDF2    & 66.97     & 53.17      & 69.05     & 84.30    & 68.37     & 30.27      \\ 
\hline
             & FF++    & 95.67 & -      & -     & -     & \textbf{95.67} & -      \\ 
\cline{2-2}
             & DFDC-P  & 93.15 & 88.87  & -     & -     & \textbf{91.01} & \textbf{2.52}      \\ 
\cline{2-2}
\textbf{Ours} & DFD     & 90.30 & 85.42  & 94.67 & -     & \textbf{90.03} & \textbf{4.41}      \\ 
\cline{2-2}
             & CDF2    & 86.28 & 79.53  & 92.36 & 83.81 & \textbf{85.49} & \textbf{7.01}      \\
\hline
\end{tabular}
\label{tab:Ab}
\end{table}
\end{small}

\subsection{Comparison of different task orders}

\begin{table}
\centering
\caption{Performance comparison of different task orders.}
\begin{tabular}{c|cccccc} 
\hline
Dataset & FF++  & DFDC-P & CDF2   & DFD   & AA    & AF    \\ 
\hline
FF++    & 95.67 & -      & -      & -     & 95.67 & -     \\ 
\cline{1-1}
DFDC-P  & 93.50 & 87.71  & -      & -     & 90.60 & 2.17  \\ 
\cline{1-1}
CDF2    & 90.02 & 84.53  & 85.26  & -     & 86.60 & 4.41  \\ 
\cline{1-1}
DFD     & 88.95 & 83.62  & 86.26  & 92.51 & 87.10 & 4.24  \\ 
\hline
Dataset & FF++  & DFD    & DFDC-P & CDF2  & AA    & AF    \\ 
\hline
FF++    & 95.67 & -      & -      & -     & 95.67 & -     \\ 
\cline{1-1}
DFD     & 91.91 & 96.54  & -      & -     & 94.23 & 3.76  \\ 
\cline{1-1}
DFDC-P  & 89.43 & 93.55  & 83.41  & -     & 88.80 & 4.61  \\ 
\cline{1-1}
CDF2    & 89.61 & 95.08  & 85.81  & 84.14 & 88.66 & 1.70  \\
\hline
\end{tabular}
\label{tab:TO}
\end{table}

 In Table \ref{tab:TO}, we provide experimental results on the effect of task order on our method. The experimental results indicate that the order of tasks affects the results a little. However, our DFIL still effectively organized catastrophic forgetting.

\section{Limitations and Future work}
In incremental learning, the strategy of replaying samples is the most common and effective method to prevent catastrophic forgetting, but there is a problem that there may be a privacy leakage problem if the original samples are stored. So we will seek to a method to protect the private content in images in DFIL.

\end{sloppypar}
\end{document}